\documentclass[runningheads]{llncs}
\usepackage[T1]{fontenc}
\usepackage{graphicx}
\usepackage{amsmath}
\usepackage{multirow}
\usepackage{tabularx}
\usepackage{multicol}
\usepackage{booktabs}
\usepackage{float}

\newcommand\MyBox[2]{
  \fbox{\lower0.75cm
    \vbox to 1.7cm{\vfil
      \hbox to 1.7cm{\hfil\parbox{1.4cm}{#1\\#2}\hfil}
      \vfil}%
  }%
}

\newcommand\MyBoxx[1]{
  \fbox{\lower0.75cm
    \vbox to 1.7cm{\vfil
      \hbox to 1.7cm{\hfil\parbox{0.46cm}{#1}\hfil}
      \vfil}%
  }%
}

\begin{document}
\title{Clustering Validation with The Area Under Precision-Recall Curves}
%
%
\author{Pablo Andretta Jaskowiak\inst{1}
\and
Ivan Gesteira Costa\inst{2}
}
\authorrunning{Jaskowiak and Costa}
%
\institute{Federal University of Santa Catarina (UFSC), Joinville, Santa Catarina, Brazil\\
\email{pablo.andretta@ufsc.br}\\
\and Institute for Computational Genomics, Joint Research Center for Computational Biomedicine, RWTH Aachen University Medical School, 52074 Aachen, Germany\\
\email{ivan.costa@rwth-aachen.de}}
\maketitle              
\begin{abstract}
Confusion matrices and derived metrics provide a comprehensive framework for the evaluation of model performance in machine learning. These are well-known and extensively employed in the supervised learning domain, particularly classification. Surprisingly, such a framework has not been fully explored in the context of clustering validation. Indeed, just recently such a gap has been bridged with the introduction of the Area Under the ROC Curve for Clustering (AUCC), an internal/relative Clustering Validation Index (CVI) that allows for clustering validation in real application scenarios. In this work we explore the Area Under Precision-Recall Curve (and related metrics) in the context of clustering validation. We show that these are not only appropriate as CVIs, but should also be preferred in the presence of cluster imbalance. We perform a comprehensive evaluation of proposed and state-of-art CVIs on real and simulated data sets. Our observations corroborate towards an unified validation framework for supervised and unsupervised learning, given that they are consistent with existing guidelines established for the evaluation of supervised learning models.


\end{abstract}
\section{Introduction}

Clustering is an unsupervised Machine Learning (ML) task, with a broad and diverse spectrum of applications. Given a data set $\mathbf{X} = \{\mathbf{x}_1, \dots, \mathbf{x}_n\}$, with $n$ objects embedded in a $m$-dimensional space, a typical clustering procedure consists in partitioning $\mathbf{X}$ into a number of clusters $k$, usually unknown beforehand. One expects that, at the end of
the clustering procedure, objects within each cluster are more similar to each other
than objects that belong to different clusters.  Henceforth we consider hard clustering solutions, i.e., partitions, with $2 \le k \le n-1$, expressed in the form $\mathcal{C} = \{C_1, \dots, C_k\}$, and subject to~(i)~${C_1} \cup \dots \cup {C_k} = \mathbf{X}$; (ii) ${C_i} \neq \emptyset, \forall i$; and (iii) ${C_i} \cap {C_j} = \emptyset, \forall i,j$  with  $i \neq j$.

Given the unsupervised nature of clustering, the vast majority of its methods will produce a result for any given set of input parameters. In such a scenario, the validation of clustering results acquires critical importance. By resorting to clustering validation procedures in the context of exploratory data analysis, one is more likely to avoid meaningless solutions, selecting only a narrow subset of relevant ones. Ideally, such solutions would be further considered and inspected by a specialist of the subject under investigation, which then might endorse and provide context to the findings. Conversely, if clustering is employed as an automated pre-processing step within a broader ML framework, e.g., feature selection, quantitative validation procedures can provide the necessary degree of automation to select the most significant clustering results that will be further considered downstream in the pipeline, without the need of human intervention.

Clustering Validity Indices (CVIs) are the ``tools'' usually applied for the quantitative validation of clustering results~\cite{Romera2019}. These mathematical indices can be broadly divided into three categories: external, internal, and relative~\cite{JaiDub88}.~External CVIs compare pairs of partitions, providing a score based on their degree of correspondence~\cite{Meila2005}. Such indices are typically employed to the development/evaluation of clustering algorithms, in which candidate clustering solutions are compared against a reference partition (ground truth) known beforehand. 
Internal validity criteria, take into account only the data under analysis and the clustering outcome (partition). Such measures are usually heavily influenced by the number of clusters from the partition under evaluation and cannot be directly applied to distinguish among results with different $k$. Finally, relative validity criteria comprehend a subset of internal CVIs that can be employed to evaluate clustering results without direct influence of their respective number of clusters~(at least ideally). Given that relative CVIs allow for the selection of ``the best'' clustering outcome (and its corresponding number of clusters), they are 
ultimately adopted in real clustering applications scenarios.

There is a significant number of relative CVIs available in the literature~\cite{VenCamHru10}, such as Silhouette, C-Index, Gamma, Dunn, and Davies-Boudin, just to mention a few. This considerable number of indices suggests that no single relative CVI can capture the notion of clustering, neither prevail in all scenarios, as particular data characteristics might affect its performance in unpredictable ways~\cite{Dubes1993}.
~Indeed, such an universal view of clustering is, at best, ill-advised, and has already been heavily criticized, with a call for the identification of classes of clustering problems that might be treated similarly~\cite{Luxburg2012}. Such an identification would not only avoid an universal and simplified view of the clustering problem, but more importantly also prevent the search for an (nonexistent) universal solution.

Just recently, it has been demonstrated that the Area Under the (ROC) Curve (AUC), previously employed mostly to the validation of binary classification models~\cite{Fawcett06,Flach2011,Flach2010} also serve as an effective and computationally efficient relative CVI~\cite{Jaskowiak2022}. In this work we further shorten the gap between the evaluation of supervised and unsupervised learning models by introducing  relative CVIs based on Area Under the Precision-Recall Curve (AUPR). Our motivation arises from the fact that AUPR has been shown to be more reliable than AUC when dealing with heavily imbalanced classification problems~\cite{DavGoa06,Saito2015,Sofaer2019}.~We, therefore, propose the adoption of AUPR based relative CVIs, targeted at the evaluation of clustering solutions with (heavy) imbalance. Such problems are far from rare~\cite{Chawla2004} and pose challenges in scenarios such as single cell data analysis~\cite{Yu2022}. 

The remainder of the paper is organized as follows. In Section~\ref{sec:related}, we provide the necessary background and revisit the Area Under the (ROC) Curve for Clustering~(AUUC), which will provide the basis for the introduction of AUPR based CVIs, in Section~\ref{sec:aupr}. In Section~\ref{sec:setup} we detail the  experimental design of our empirical evaluation procedure, for which results are given in Section~\ref{sec:results}. Final remarks and possible future work directions are discussed in Section~\ref{sec:conclusions}.

\section{Background}\label{sec:related}

A relative CVI (Cluster Validity Index) allows for the quantitative evaluation of a clustering solution. It provides a numerical score that relates to the quality of the partition under evaluation, according, of course, to that specific CVI. In order to select the most suited clustering solution among a set of candidate ones (generated, for instance, with different clustering algorithms and/or parameters), a user can apply a relative CVI and select the partition that yields the best score. 

Relative CVIs usually consider two basic characteristics of a partition to quantify its overall quality, namely, intra-cluster compactness and inter cluster separation~\cite{XuWun09}. A number of indices have been proposed based on different definitions/formulations of these two concepts. A handful of reviews and empirical evaluations of relative CVIs can be found in the literature~\cite{Mil81,MilCoo85,Maulik02,VenCamHru10,Arbelaitz2013}.~Due to their practical appeal, implementations for a number of such indices are readily available, e.g., in Python and R Packages~\cite{clValid2008,NbClust,ClusterCrit2016,scikit-learn}.




Just recently the Area Under the (ROC) Curve (AUC) has been borrowed from the supervised learning domain and considered as a relative CVI for the validation of clustering results~\cite{Jaskowiak2022}. The Area Under the (ROC) Curve of a clustering solution which, in order to differentiate from AUC in the supervised setting,  is referred to as Area Under the (ROC) Curve for Clustering (AUCC), can be computed based on: (i)~a pairwise clustering solution and; (ii) the pairwise similarities from the objects that compose such solution. Given a clustering solution, i.e., a partition $\mathcal{C}$, a pairwise clustering solution $\mathcal{C}^p$, with a total of $n(n-1)/2$ distinct pairs, can be readily obtained with the following rule:

\begin{displaymath}
\label{adap2}
\mathcal{C}^p(\mathbf{x}_i,\mathbf{x}_j) = \left\{ 
\begin{array}{l l}
  1 & \quad \mbox{if } \exists l: \mathbf{x}_i,\mathbf{x}_j \in C_l,\\ \noalign{\medskip}
  0 & \quad \mbox{otherwise}.\\
\end{array} \right.
\end{displaymath}

\noindent Pairwise similarities can be readily obtained by the transformation of distance metrics, i.e., Euclidean distance, into similarities. 

Once the two pairwise relationships previously discussed are available, they can be provided as input to a standard ROC Analysis procedure~\cite{Fawcett06,Flach2011,Flach2010}, which, in the context of clustering validation, can be briefly summarized as follows. Similarity values between all object pairs are sorted in a decreasing fashion.~Each numerically distinct similarity value is considered as a decision threshold~$\phi_s$, which establishes that: object pairs with a similarity higher than $\phi_s$ are deemed ``positive''~(should be in a same cluster), otherwise, they are deemed ``negative'' (should be in different clusters).~Binary predictions, as derived from a given similarity threshold~$\phi_s$, can be compared against the ones established by the pairwise clustering solution $\mathcal{C}^p$. With the aid of a confusion matrix (contingency table), as depicted by Figure~\ref{fig:confusion_matrix}, each pair of objects can, on the basis of their pairwise similarity/threshold and clustering membership, be regarded as: True Positive (TP), False Positive (FP), False Negative (FN), or True Negative (TN). Finally, a ROC Plot can be obtained by relating all the values of two particular metrics derived from the confusion matrices, namely the True Positive rate ($TPR$) and false negatie rate ($FPR$). From such a ROC Curve, the corresponding Area Under the Curve for Clustering (AUCC) can be obtained, thus providing the quality score for the partition at hand.

\begin{figure}[htpb]
  \centering
  \begin{minipage}{0.39\linewidth}
  {\begin{center}
  \renewcommand\arraystretch{1.5}
  \setlength\tabcolsep{2pt}
  \begin{tabular}{c >{\bfseries}r @{\hspace{0.7em}}c @{\hspace{0.4em}}c @{\hspace{0.7em}}c}
  \multirow{9}{*}{\rotatebox{90}{\parbox{6.2cm}{\bfseries\centering Prediction ($\phi_S$)}}} & 
  & \multicolumn{2}{c}{\bfseries \hspace*{-0.3cm}Expected ($\mathcal{C}^p$)} & \\\addlinespace[-0.1cm]
  & & \bfseries Positive & \bfseries Negative & \bfseries \\
  & \rotatebox[origin=c]{90}{\parbox{1.6cm}{\bfseries\centering Positive}} & \MyBoxx{$TP$} & \MyBoxx{$FP$} &\\[25pt]
  & \rotatebox[origin=c]{90}{\parbox{1.6cm}{\bfseries\centering Negative}}  & \MyBoxx{$FN$} & \MyBoxx{$TN$} &\\
  \end{tabular}
  \end{center}}
  \end{minipage}\hspace{1.2cm}
  \begin{minipage}{0.5\linewidth}

  \underline{\textbf{AUC/AUCC Metrics}}
  
  \vspace{0.2cm}
  $TPR = TP/(TP+FP)$

  \vspace{0.2cm}
  $FPR = FP/(FP+TN)$

  \vspace{1cm}
  \underline{\textbf{AUPRC/AUPRCC/Related Metrics}}
    
  \vspace{0.2cm}
  $\text{Precision} = TP/(TP+FP)$

  \vspace{0.2cm}
  $\text{Recall} = TP/(TP+FN)$

  \vspace{0.2cm}
  $\text{Invese Precision} = TN / (TN + FN)$  

  \vspace{0.2cm}
  $\text{Invese Recall} = TN / (TN+FP)$
  \end{minipage}
  
  \caption{A confusion matrix/contingency table (left) and some metrics (right) that can be readily extracted from it for computing AUC~(Area Under the ROC Curve),~AUPR~(Area Under Precision-Recall curve) and related metrics. Note that Inverse Precision and Inverse Recall are also known as Negative Predictive Value and True Negative Rate, respectively. We opt for the ``Inverse'' terminology from Powers~\cite{Powers2011}, given that it emphasize the fact that these are the values of Precision and Recall w.r.t. the negative class (Powers refers to this as the ``Inverse Problem''), in which one assumes that the negative class is that of prevalent interest, contrary to the positive one.}
  \label{fig:confusion_matrix}
  \end{figure}

The rationale behind this evaluation is that in a good clustering solution, object pairs in a same cluster should have higher similarities than those in different clusters. The relationship to ROC Analysis in classification is straightforward: similarity values correspond to ``classification thresholds'' whereas pairwise clustering memberships correspond to the ``true classes''. It has been shown that AUCC holds an expected value of 0.5 for random clustering solutions and, moreover, that it is actually a linear transformation of the Gamma relative validity index from Baker and Hubert~\cite{BakHub75}, but with a much lower computational cost~\cite{Jaskowiak2022}.

\section{The Area Under Precision-Recall Curve for Clustering}\label{sec:aupr}

Even though results for AUCC are promising for relative clustering validation in a general context, it is important to keep in mind that the ROC Curve and its corresponding AUC are not the only quality measures that can be derived from confusion matrices. In fact, in the context of supervised learning, AUC is already known to perform poorly as a metric of quality for heavily imbalanced problems~\cite{DavGoa06,Saito2015,Sofaer2019}. These are our main motivations to explore AUPR based metrics in the context of relative clustering validation. Bearing that in mind, a counterpart of the Area Under Precision-Recall curve (AUPR) for the relative validation of clustering results can be derived similarly as shown before for AUCC. The major change arises from the fact that instead of relating values of $TPR$ and $FPR$ for each one of the confusion matrices we obtain during the process (one for each different similarity value), we now relate values of Precision and Recall, which can easily computed as shown in Figure~\ref{fig:confusion_matrix}. 

It is worth noticing that in the context of binary classification problems (supervised learning), the positive class usually has a somewhat greater importance than the negative one. This is due to the fact that the positive class is generally associated with the ``case'' of interest, which one aims to correctly identify.~Take, for instance, the problem of predicting whether a patient is sick~(positive) or healthy (negative). It can be argued that the misclassification of a sick patient as healthy is more severe than the misclassification of a healthy patient as sick.~Therefore, in such problems, most of the interest lies in: (i) the success rate of the classifier w.r.t. its positive predictions, that is, from its positive predictions, how many are actually positive (precision) and; (ii) the success rate of the classifier in recovering/identifying  actual positive cases (recall).

In some problems, however, such asymmetry might simply not exist. In other problems, it may present itself reversed. In the latter case, the negative class has more ``importance'' than the positive one,  as discussed by Powers~\cite{Powers2011}, which refers to this as ``Inverse Problem''. By taking these scenarios into account, one may define counterpart metrics of Precision and Recall that provide evaluation scores on the basis of the negative class, namely, Inverse Precision (Negative Predicted Value) and Inverse Recall (True Negative Rate); please, refer to Figure~\ref{fig:confusion_matrix}. Indeed, such an observation has already led to the combination of measures/metrics and their inverse counterparts in the context of external validation of clustering results, see for instance~\cite{AmiGonArtVer09}.

In the context of relative clustering validation, the problem we address in the current paper, such an asymmetry between positive and negative classes is neither observed nor desired. Notice, from the definition of AUCC, that one deals with clustering memberships and similarities in terms of object pairs. Therefore, all confusion matrices and statistics derived from them are also w.r.t. object pairs.~In this context, assigning a pair of objects from a same cluster (positive pair) to different clusters (negative pair) is as bad as assigning a pair of objects from different clusters (negative pair) to a same cluster (positive pair). 

Therefore, we consider for evaluation three relative CVIs based on the concepts of the Area Under Precision-Recall curve (AUPR) and its Inverse counterparts, namely: (i) the Area Under Precision-Recall curve for Clustering (AUPRC), which relates the values of Precision and Recall; (ii) the Area Under Inverse Precision-Recall curve for Clustering (AUIPRC), which relates the values of Inverse Precision and Inverse Recall; and (iii) the average between AUPRC and AUIPRC, which we refer to as Symmetric Area Under Precision-Recall curves Clustering (SAUPRC). Note that the actual procedure adopted in order to obtain such metrics is analogous to the one previously discussed in the context of AUCC, except for the metrics which are derived from the confusion matrices.

\section{Experimental Design}\label{sec:setup}

\subsection{Evaluation Procedure}

In order to empirically assess the performance of the relative CVIs considered in this study, we resort to a well established evaluation procedure from the literature. This procedure was first discussed by \cite{Mil81} and more recently formalized and employed by \cite{VenCamHru10} to the empirical evaluation of a number of relative CVIs. In brief, it is based on the reasoning that a good relative CVI should provide evaluations that agree with that of an external CVI. Given a data set with known cluster structure, i.e., a ground truth, the evaluation procedure is as follows:

\begin{enumerate}
  \item Generate partitions of varied qualities for the data set.~This may be achieved by considering different clustering algorithms and configurations of their parameters, ideally, leading to partitions with different numbers of clusters ($k$);
  \item For each one of the partitions from the previous step, compute its quality according to: (i) the relative CVI  under evaluation and; (ii) an external CVI that quantifies  its agreement w.r.t. the ground truth partition;
  \item Correlate the evaluation scores of each partition, as obtained from the relative and external evaluations. This indicates the quality of the relative CVI.
\end{enumerate}

Four definitions are needed within this evaluation framework. The first one is that of the clustering algorithms employed in order to generate a varied set of partitions. In our study we considered k-means and agglomerative hierarchical clustering algorithms with the following linkage functions: Single, Average, Complete, and Ward's~\cite{XuWun09}. The second one corresponds to the range of the number of clusters ($k$) allowed for the partitions, for which we consider $2 \le k \le \lceil \sqrt{n} \rceil$, where $n$ is the number of objects from the data set.
~The third and fourth definitions are related to the external index and the correlation between external and relative evaluations. For those we consider two well-known measures, i.e., the Adjusted Rand Index (ARI)~\cite{Hubert1985} and the Pearson correlation coefficient~\cite{Freedman98}. 

We evaluate within the same experimental setting several well established relative CVIs from the literature~\cite{Nguyen2020,Zhou2021}. These are the same\footnote{In the evaluation of AUCC~\cite{Jaskowiak2022}, in addition to their original formulations, 3 variants of SWC and 17 variants of Dunn were considered. Given that such variants provided similar results to their ``original'' counterparts, we believe that their inclusion would bring more distractions than insights and, therefore, do not consider them here. } measures employed in the previous study that introduced AUCC~\cite{Jaskowiak2022}. In brief, besides the three relative CVIs we introduce in this work, i.e., AUPRC, AUIPRC, and SAUPRC, we also take into account~AUCC, Silhouette Width Criterion (SWC) \cite{Rou87}, Davies-Bouldin (DB) \cite{Davies1979}, C Index \cite{Hubert1976}, Dunn \cite{Dunn1974}, PBM \cite{PakBanMau04}, Calinski-Harabasz (VRC) \cite{Calinski74}, Point Biserial (PB) \cite{Mil81} and, Ratkowsky Lance (C/Sqrt(k)~\cite{RatLan78}. For a thorough review of such measures we refer the reader to 
 Vendramin et. al~\cite{VenCamHru10}.

\subsection{Evaluation Scenarios}

Three data sources are considered in the evaluation, as we detail in the following.
\subsubsection{Julia Handl} This set comprises a benchmark of 80 data sets from~\cite{Handl2007}, which account for multivariate normally distributed clusters. The number of objects within each data set ranges from around 800 to 4.000~(almost all datasets have different numbers of objects). Given that this is a synthetic benchmark set, the following parameters were employed by the aforementioned authors to generate them. Regarding dimensions, data sets have 2 or 20 dimensions. The numbers of clusters ($k$) considered are 4, 10, 20, and 40. Two different cluster sizes are considered by authors, depending on the number of clusters present in the data set. For data sets with 4 and 10 clusters, their sizes are uniformly drawn within [50,500]; whereas for data sets with 20 and 40 clusters, their sizes are uniformly drawn within [10,100].~Data sets were generated iteratively, preventing overlap between clusters, in a trial and error procedure. For more details regarding the data generation procedure, the reader is referred to~\cite{Handl2007}. A detailed description of the generator and the actual data sets are provided in one of the author's webpage: \url{https://personalpages.manchester.ac.uk/staff/julia.handl/}.

\subsubsection{Tabula Muris} This set of data is comprised of 64 data sets from the Tabula Muris Compendium~\cite{Schaum2018}, which is based on single cell experiments from different organs and tissues of the \emph{Mus musculus} organism. The following 20 organs/tissues are considered: aorta, bladder, brain~(myeloid), brain (non-myeloid), diaphragm, fat, heart, kidney, large intestine, limb muscle, liver, lung, mammary gland, marrow, pancreas, skin, spleen, thymus, tongue and, trachea. Data from the two available single cell isolation methods (FACS and droplet) were considered. Previously normalized data were used~\cite{Schaum2018}. In order to obtain the ground truth for each data set, allowing thus for the use of an external index, we considered the two available options: cluster IDs and cell ontology class. Data sets were obtained with the Seurat Package~\cite{Seurat}.

\subsubsection{Synthetic Data sets} This set of 2D Synthetic data was generated for this study, in an iterative fashion. Cluster centers were constrained within the~[0,500] interval in both $x$ and $y$ axis. Cluster variances for both axis were identical, randomly drawn from the the [2,50] interval. For a given data set, each cluster center was randomly drawn in a sequential manner. If the newly generated cluster did not overlap with any of the existing clusters (considering, their respective variances), it was thus added to the data set. The process then continued until all the desired clusters could be added to the data set. If a new cluster could no be added to the data set after 10,000 attempts (center draws), the maximum variance was decreased in 5\% and the process would continue, with an attempt to add a new cluster. Subsequent 5\% decreases could occur (each after another round of 10,000 failed attempts). It is wort noticing, however, that this reduction was only necessary in a few cases during the generation of data sets with large numbers of clusters. In total, 2,100 data sets were generated, from the following configurations: (i) number of objects fixed in 500; (ii) 21 different numbers of clusters, in the $[2,22]$ interval; (iii) 10 different cluster imbalances within the $[0, 10, 20,\dots,80, 90]$ interval, where each value corresponds to a percentage of the number of objects assigned to one of the clusters, with the remaining ones having the same size --- 0(\%) accounts for no imbalance. Combining number of clusters and imbalances, we had a total of 210 configurations, for which we generated 10 different data sets (replicates), thus totalling 2,100. 

\subsubsection{Data Characteristics}

\begin{figure}[H]
\centering
  \includegraphics[width=0.9\linewidth,trim=0.2cm 0.2cm 0.3cm 0.4cm]
  {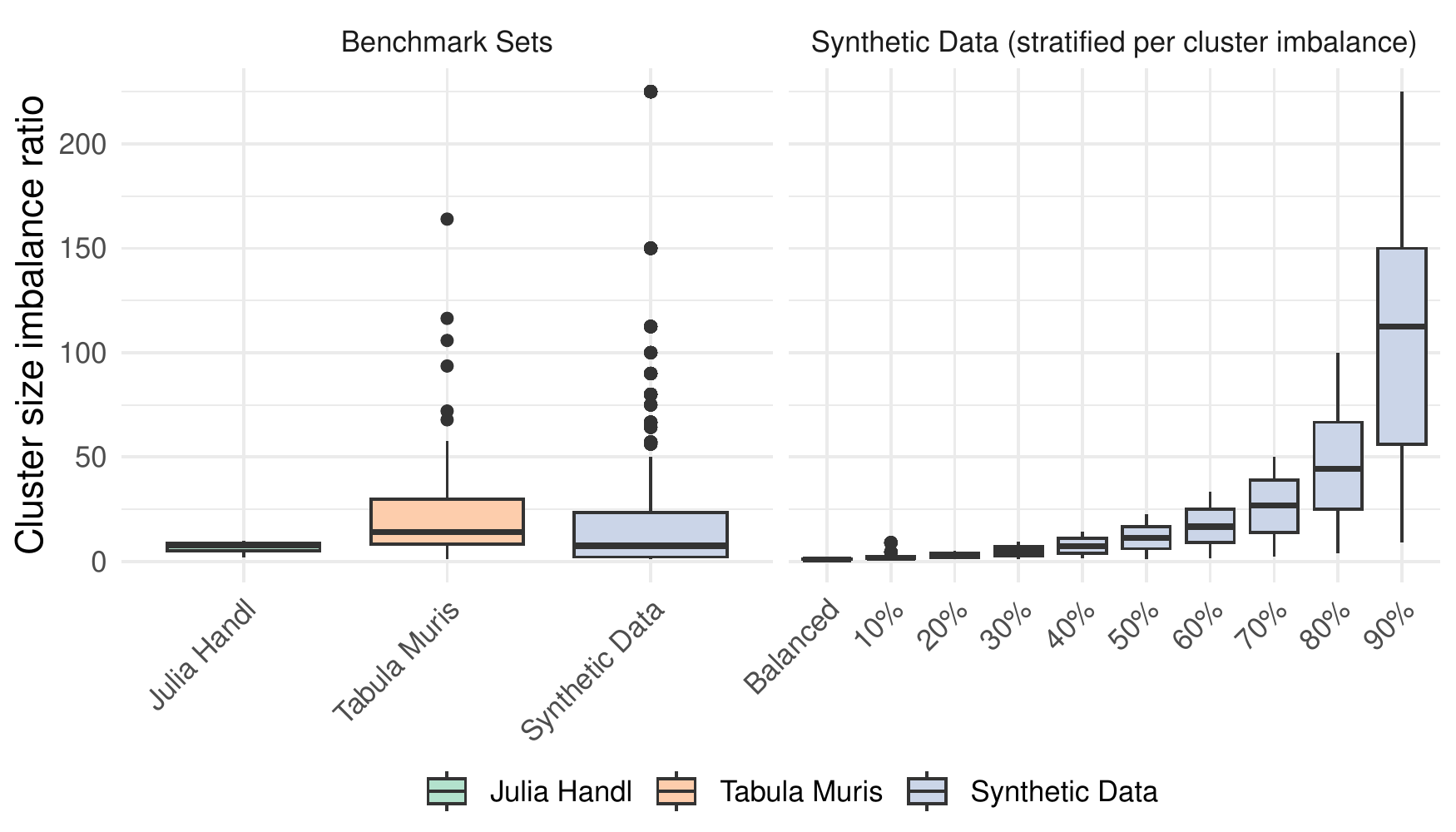}
  \caption{Characterization of cluster size imbalance in the distinct evaluation scenarios.}
  \label{fig:ratio}
\end{figure}

In order to provide some notion of the cluster imbalance present in these data sets we depict, in Figure~\ref{fig:ratio}, boxplots regarding the cluster size imbalance ratio (i.e., largest cluster size divided by smallest cluster size) for each benchmark set (left), as well as a stratified analysis per imbalance for the Synthetic data (right). Julia Handl data sets are, in general, well balanced. As for Tabula Muris, a higher imbalance can be observed, with some extreme cases, in which the largest cluster is approximately $150$ times larger than the smallest one. For the Synthetic data, we have a wide range of cluster size imbalance ratios, as defined by their design.~These diverse data sets will allow us to evaluate CVIs under distinct conditions, from completely balanced to heavy imbalance. 

\section{Results}\label{sec:results}

\subsection{Overview}

\begin{figure}
\includegraphics[width=\linewidth]{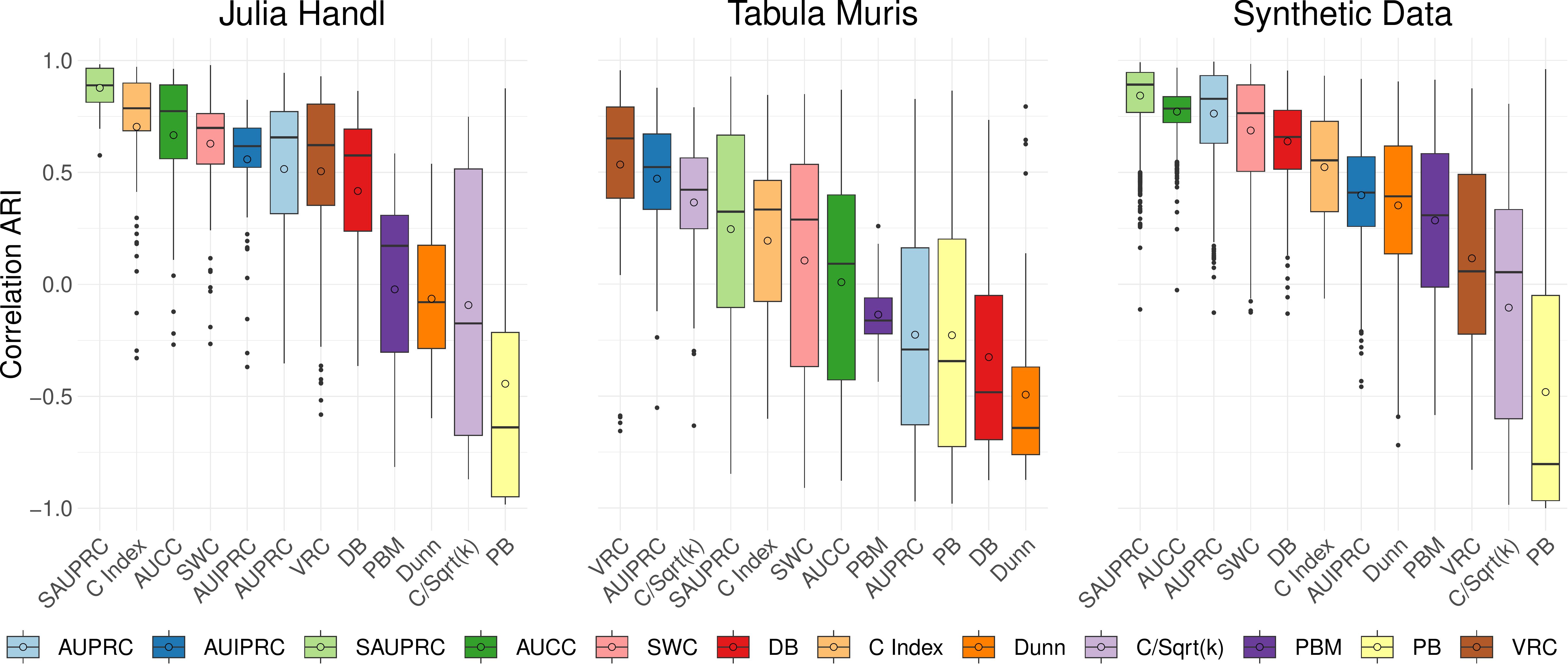}
\caption{Boxplots depict correlation between the evaluations of each relative CVI and those of the Adjusted Rand Index (ARI), i.e., the external CVI. Results are sorted according to the average correlation value, which is indicated by a circle in each boxplot.}
  \label{fig:resall}
\end{figure}

Due to the distinct characteristics of each set of benchmark data, we investigate them individually (Figure~\ref{fig:resall}). We observe that Symmetric AUPR (SAUPRC) obtains the highest correlation w.r.t. external evaluations from ARI for both Julia Handl and Synthetic data. In the case of the Tabula Muris benchmark set, VRC (Calinski Harabasz) provided the overall best results, in terms of median correlation with ARI, followed by AUIPRC. Nonetheless, SAUPRC appears as the fourth best CVI for this benchmark set. In general terms, SAUPRC provides more consistent/robust evaluations than its base counterparts, which are also evaluated here, namely, AUPRC and AUIPRC. Except for Julia Handl data sets, for which AUPRC and AUIPRC appear next to each other, results for these measure are quite contrasting for Tabula Muris and Synthetic data, which can be observed by the change in their relative order w.r.t. the boxplots from Figure~\ref{fig:resall}. Taking such behaviours into account, by combining AUPR and AUIPR, SAUPR provides robust evaluations in different settings.


It is worth noticing that in all benchmark sets SAUPRC provided better results than SWC (Silhouette), which is one of the most commonly employed CVIs from the clustering literature. Moreover, results for SAUPRC are superior than those of the recently introduced AUCC in all benchmark sets. Altogether, these results indicate the power of CVIs based on metrics from confusion matrices, given that AUCC also shows competitive results to SWC in all settings. 

\begin{figure}
\centering
  \includegraphics[width=0.32\linewidth,trim=0.9cm 0.6cm 0.9cm 0.4cm]{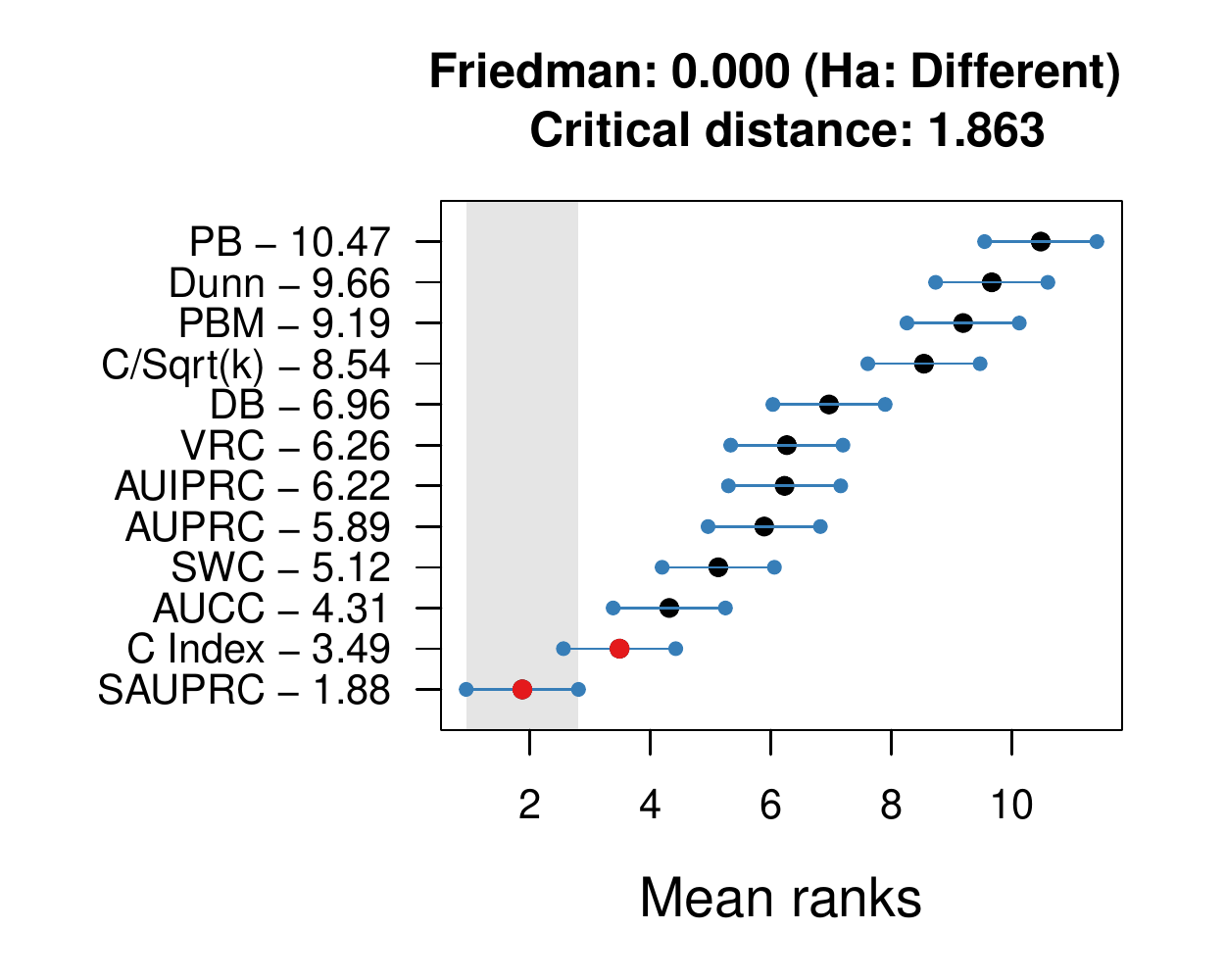}\hfill
  \includegraphics[width=0.32\linewidth,trim=0.9cm 0.6cm 0.9cm 0.4cm]{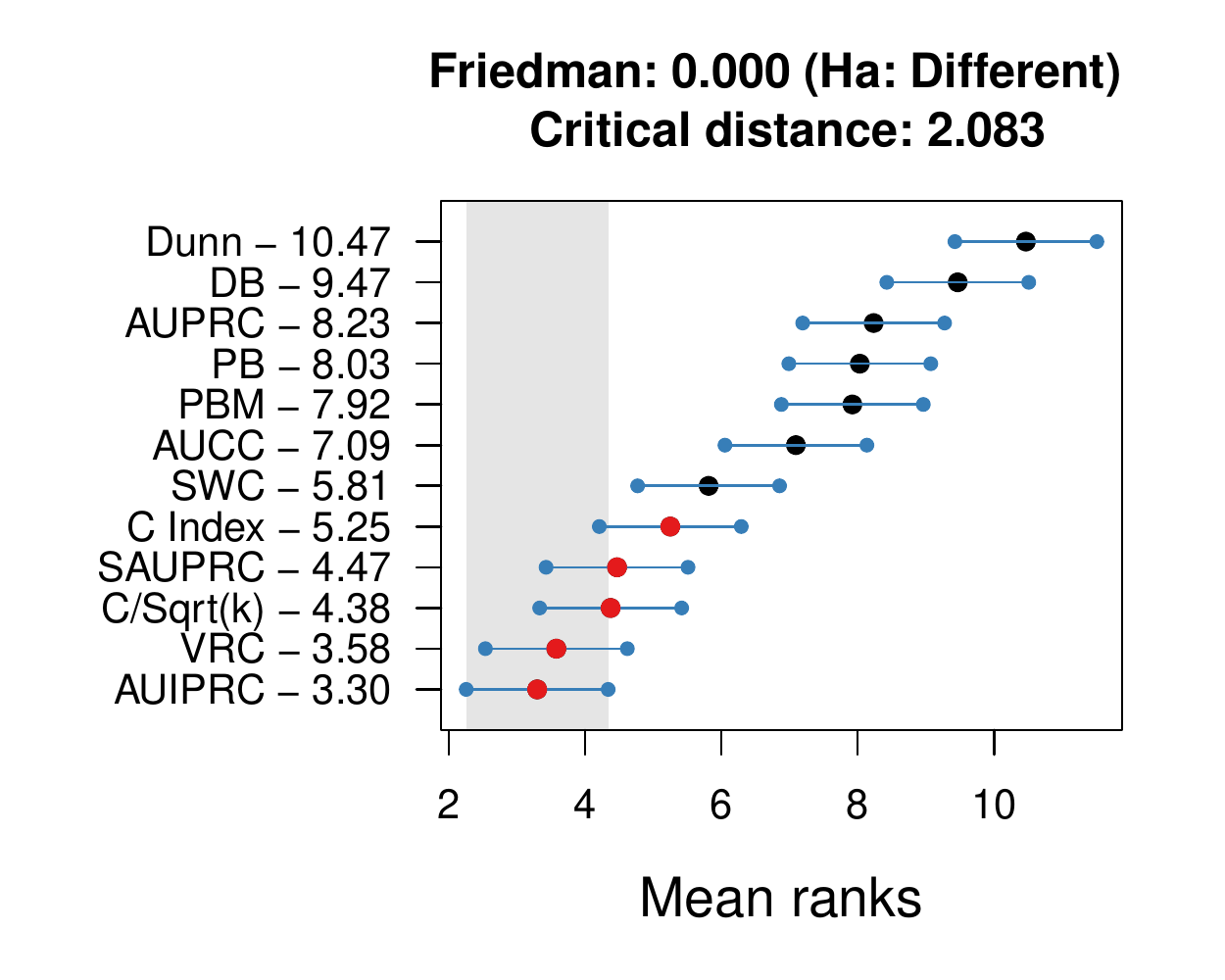}\hfill  
  \includegraphics[width=0.32\linewidth,trim=0.9cm 0.6cm 0.9cm 0.4cm]{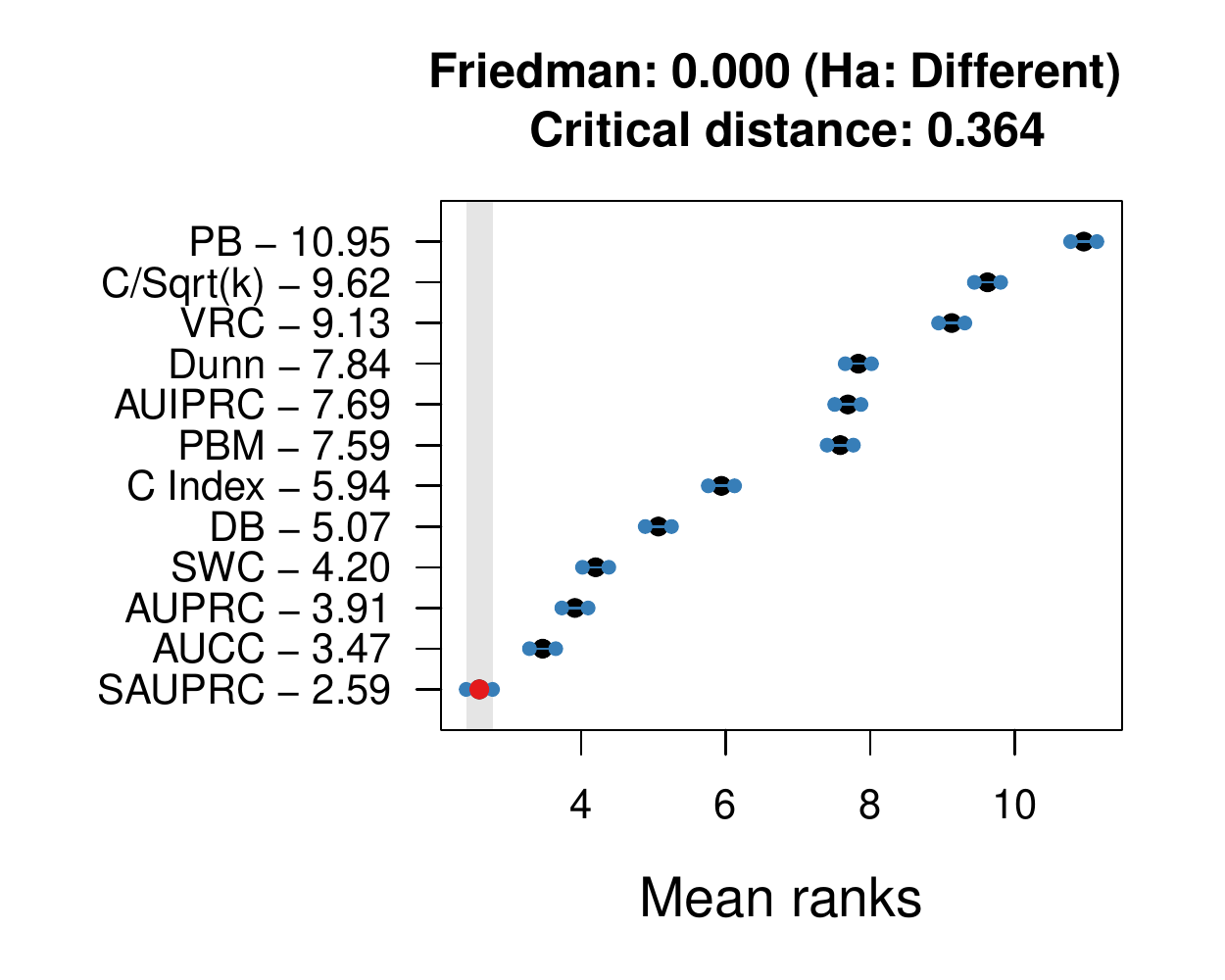}
  \caption{Results regarding Friedman-Nemenyi statistical tests for the three benchmark sets: Julia Handl (left), Tabula Muris (center) and Synthetic data (right). Mean rank values for each method are shown in the vertical axis (lower mean ranks indicate better methods). The horizontal line for each method accounts for its mean rank (central dot), as well as the critical difference established by the test (highlighted in gray). A statistically significant difference is observed between CVIs (in favor of the lower ranked one) if the projections of their horizontal lines do not overlap in the $x$ axis.
  }
  \label{fig:testall}
\end{figure}

We applied statistical tests taking into account the results obtained within each benchmark set. Given that we are dealing with the comparison of multiple methods (CVIs) in multiple data sets, we resorted to the Friedman statistical test followed by the Nemenyi posthoc test~\cite{Dem06}. Results are shown in Figure~\ref{fig:testall}. Apart from the statistical differences themselves, these plots provide a nice summary w.r.t. the mean ranks of each CVI. In terms of mean ranks, SAUPRC is the best CVI for both Julia Handl and Synthetic data. As for Tabula Muris, it ranks~4.47. In this latter set, VRC has a slightly higher mean rank than AUIPRC (the lower the rank, the better the method). In terms of statistical differences, the overall scenario considering each benchmark set is the following: (i) for Julia Handl, SAUPRC is statistically superior than all the remaining CVIs, except for C Index; (ii) for Tabula Muris, VRC and AUIPRC provide the top results, but are not statistically superior than C/Sqrt(k), SAUPRC, and C Index; and (iii) for the Syntetic set, SAUPRC is superiror to all the remaining CVIs.

\subsection{Cluster Size Imbalance Effect}

One advantage of simulated data is that we can check how different cluster imbalances affect the performance of the CVIs. Here we present results for the Synthetic data in a stratified manner, considering each imbalance separately. As previously discussed, 10 different cluster imbalances within the $[0, 10, 20,\dots,80, 90]$ interval were considered. Each imbalance value corresponds to the percentage of objects assigned to a single cluster,  with the remaining objects uniformly distributed among remaining clusters. The value 0(\%) accounts for no imbalance. 

\begin{figure}[H]
  \includegraphics[width=\linewidth]{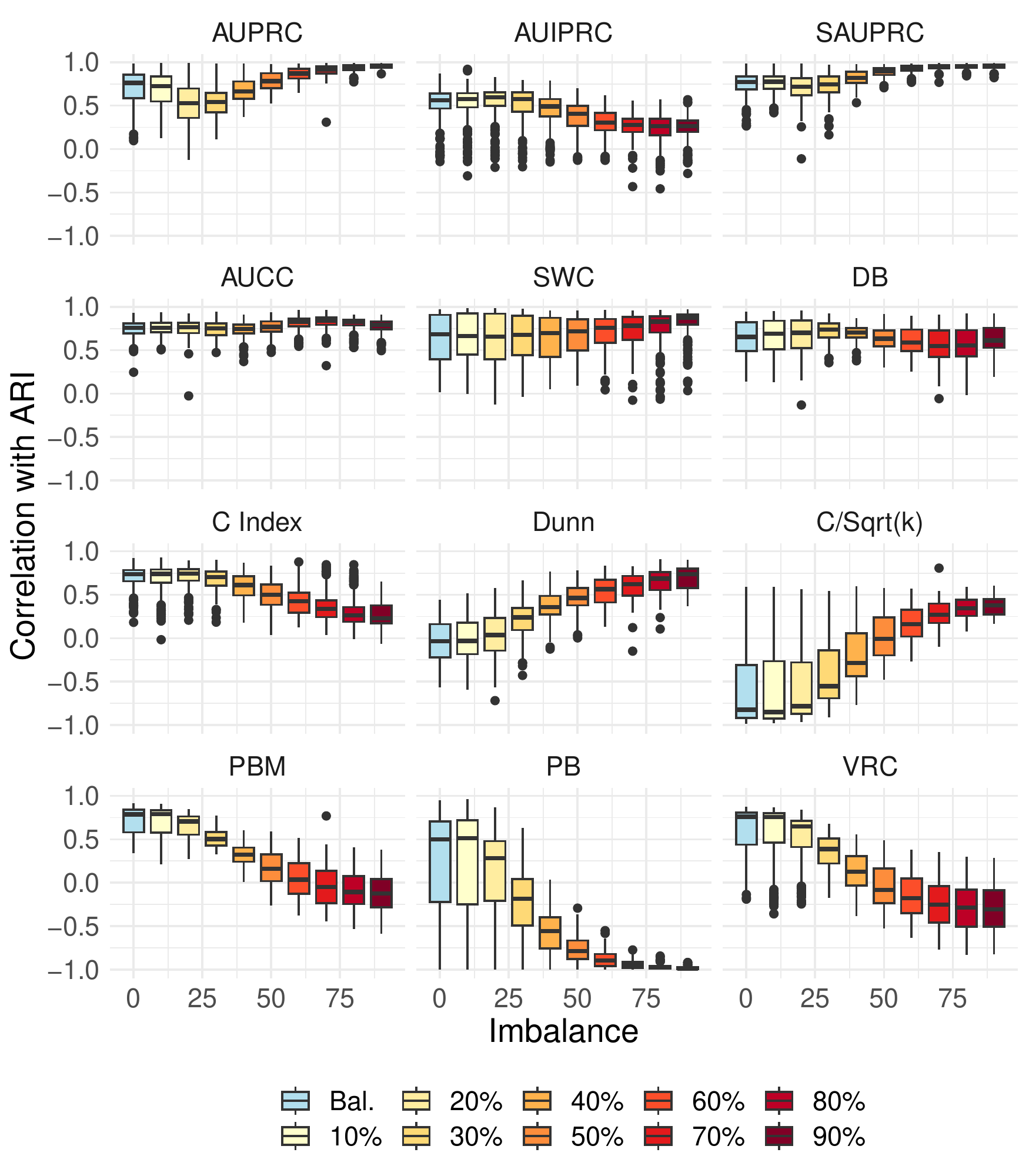}
  \caption{Correlation of relative CVIs with external index (ARI) considering the different cluster imbalances from the Synthetic data sets. Balanced data is indicated by ``Bal.''. }
  \label{fig:imb}
\end{figure}

Results for such analysis are depicted by Fig.~\ref{fig:imb}. Broadly speaking, three distinct performance behaviours can be observed from the plot w.r.t. imbalance.

The first one is that of CVIs with  deteriorating performance as cluster imbalance increases. This trend is noticeable for AUIPRC, C Index, PBM, PB, and VRC. A quite extreme case is that 
of PB (Point Biserial), for which the median correlation transitions from values close to $0.5$ in the case of balanced data to~$-1.0$ in the highest imbalance considered in our evaluation design ($90\%$). 

The second distinctive behaviour is that of measures with a (mostly) stable performance w.r.t. class imbalance. Within this category we can place AUCC, DB, and SWC (arguably).~In the case of SWC, performance is actually slightly better with increasing imbalance. It is worth noticing, however, that this particular CVI provides a high number of outlying evaluations for higher imbalances.

Finally, the third observable behavior is that of CVIs that provide better evaluations as cluster imbalance increases. In this category we can place AUPRC, SAUPRC, Dunn, and C/Sqrt(k). The case of C/Sqrt(k) is quite peculiar, as the evaluations of this particular CVI provide a negative agreement w.r.t. those of the external index with balanced data. Its performance improves as imbalance rises, but median correlation values stay below $0.5$, indicating only a moderate agreement w.r.t. external evaluations. In the case of Dunn, a similar analysis is possible. Its median correlations with ARI range, however, from $0.0$ (no imbalance) to $0.75$ ($90\%$ imbalance). Here, SAUPRC and AUPRC stand out as the superior CVIs within this category. Notice that even though their performance is enhanced as imbalance increases, reaching median correlation values close to $1.0$, for balanced data, their median correlation values are already above $0.75$. Finally, it is interesting to observe that SAUPRC combined the best behaviours of AUIPRC, which has its best performance for balanced data, and  AUPRC, for which best performance is observed with high imbalance data.


\section{Conclusions}\label{sec:conclusions}

The positive results obtained in the context of relative clustering validation with the Area Under the ROC Curve for Clustering~(AUCC)~\cite{Jaskowiak2022} opened new venues of investigation. In the context of supervised learning, AUC is just one among several metrics that can be derived from confusion matrices in order to quantify the performance of classification models. In the realm of relative clustering validation, however, most of this measures remain yet unexplored.~In this work, we proposed and investigated empirically three CVIs based on the concept of the Area Under Precision-Recall (AUPR), taking a step further towards a unified framework for the evaluation of supervised and unsupervised models.

Our evaluation was conducted on synthetic data sets well established in the clustering literature, i.e., the Julia Handl benchmark set, and data sets with different cluster size imbalances, introduced by us, in order to isolate and evaluate cluster size imbalance effect. We also considered data sets of  Single Cell experiments from the Tabula Muris project, which consists of a real application scenario with different levels of imbalances in cluster sizes. 

In general terms, the Symetric Area Under Precision-Recall for Clustering~(SAUPRC), which combines the evaluations obtained from AUPRC and AUIPRC provided the best results for cases of high cluster size imbalance. This particular measure also provided superior results for well balanced data, with better overall results than established CVIs from the literature, such as Silhouette (SWC) or AUCC. For such a reason, we believe that SAUPRC should be favored, particularly in the presence of heavy cluster size imbalance. 

As future work, we aim to investigate the effect of sub sampling in the performance of metrics based on a given area under the curve (either ROC or Precision-Recall Curves). Given that to compute such measures for a data set with $n$ objects we need to derive a pairwise clustering solution of size $n(n-1)$, it is possible that a sub sampling strategy may reduce the computational time needed in order to compute such measure, without deteriorating their performance. The actual degree of possible reduction, however, needs to be investigated.

\section*{Acknowledgements}


PAJ thanks the German Academic Exchange Service (DAAD) for financial support under the Research Stays for University Academics and Scientists 2022 Program, allowing him to conduct this research at RWTH Aachen University. 
%
%
%
\bibliographystyle{splncs04}
\bibliography{refs}
\end{document}